# Decision Making with Interval Influence Diagrams

John S. Breese
breese@rpal.com

Kenneth W. Fertig
fertig@rpal.com

Rockwell Science Center, Palo Alto Laboratory
444 High Street
Palo Alto, CA 94301

## Abstract

In previous work [1, 2] we defined a mechanism for performing probabilistic reasoning in influence diagrams using interval rather than point-valued probabilities. In this paper we extend these procedures to incorporate decision nodes and interval-valued value functions in the diagram. We derive the procedures for chance node removal (calculating expected value) and decision node removal (optimization) in influence diagrams where lower bounds on probabilities are stored at each chance node and interval bounds are stored on the value function associated with the diagram's value node. The output of the algorithm are a set of admissible alternatives for each decision variable and a set of bounds on expected value based on the imprecision in the input. The procedure can be viewed as an approximtion to a full $n$-dimensional sensitivity analysis where $n$ are the number of imprecise probability distributions in the input. We show the transformations are optimal and sound. The performance of the algorithm on an influence diagrams is investigated and comparerd to an exact algorithm.

## 1 Introduction

The difficulty and expense of assessing probabilities for large models has motivated research in techniques for perform reasoning under uncertainty with underspecified or constraints on probabilities. Our work in this area [1, 2] developed a language of independent lower bounds on component probabilities in a belief network as a means of expressing the imprecision in probabilities. In this paper we extend the previous analysis to include influence diagrams which contain decision and value nodes.

This extension provides a capability for assessing the robustness of a set of decision recommendations from an influence diagram given imprecision in probability and utility assessments. Exact verification of the sensitivity of recommendations to all possible combinations of imprecise inputs is extremely costly from a computational perspective. The procedure developed here reduces the computational cost to that of solving an influence diagram once. In Section 5 we explore the nature of this approximation relative to an exact procedure.

An influence diagram $I = \langle N, A \rangle$ consists of a set of nodes $N$ and arcs $A$. The set of nodes $N = \mathcal{U} \cup \mathcal{D} \cup \{V\}$, where $\mathcal{U}$ is a set of chance nodes, $\mathcal{D}$ is a set of decision nodes, and $V$ is the single value node. Associated with each node $X \in \mathcal{U}$ is a set of conditional probability distributions relating $X$'s outcomes to those of its predecessors ($\Pi_X$) in the graph. Interval influence diagrams differ from the standard influence diagram formalism in that we specify lower bounds on the probability distributions associated with each chance node. The lower bounds are interpreted as follows: we admit any probability interpretation, $p$, for the diagram iff

$$\forall X \in \mathcal{U}, \ b(x|\pi_X) \leq p(x|\pi_X),$$

where $\pi_X$ is an outcome of the combined set of states of the predecessors of $X$. Lower bounds for the probability of each possible value of the node given its predecessors are defined for all chance nodes in the graph. The upper bound $u(x|\pi_X)$ on each probability is implicit in the lower bounds:

$$u(x|\pi_X) = 1 - \sum_{x' \neq x} b(x'|\pi_X)$$

We have defined operations of chance node removal (corresponding to marginalization) and arc reversal (corresponding to an application of Bayes' rule) when uncertainty is expressed in terms of lower bounds and conditional independence is captured by the topology of the influence diagram [1, 2]. In this paper we define operations of chance node removal into the value node (corresponding to taking a conditional expected value) and decision node removal (corresponding to maximizing expected value). The expected value is expressed in terms of a lower and upper bound, expressing the imprecision in value for each combination of predecessors. Processing of decision nodes will generate sets of decision alternatives which are admissible based on the imprecision in the input probabilities and values, in a manner analogous to the analysis of sets of distributions consistent with a model developed previously.

## 2 Definitions

Let $p_\mathcal{U}$ denote a probability distribution over the space of variables represented by the nodes in $\mathcal{U}$, conditioned on each possible alternate decision set expressible by the nodes in D. Let $\mathcal{P}_\mathcal{U}$ denote the class of all such distributions. Similarly, let $\mathcal{V}$ denote the set of all value functions for a value node $V$ given its predecessors.[1]

**Definition 1 (Value Function Set)** *The set,*

$$\mathcal{V}_V = \{v : \Pi_V \to \Re\}$$

*is the* value function set *associated with a value node $V$ in $I$. We will drop the subscript on $\mathcal{V}_V$ when the value node designation is obvious.*

A general constraint is any subset of the sets $\mathcal{P}$ or $\mathcal{V}$. Thus, $c \subseteq \mathcal{P}$ is a general constraint on distributions in $\mathcal{P}$, and $\nu \subseteq \mathcal{V}$ is a general constraint on value functions in $\mathcal{V}$. The *regular extension* of a probability constraint was defined in [1, 2] as follows:

**Definition 2 (Regular Extension, Probability)** *The set $c^*$ is the* regular extension *of a constraint $c$ iff*

$$c^* = \{p \in \mathcal{P} | p(x) \geq \inf_{p' \in c} p'(x)\}.$$

Similarly, we now define the *regular extension* of a value constraint, including upper and lower limits explicitly:

**Definition 3 (Regular Extension, Value)** *The set $\nu^*$ is the* regular extension *of $\nu$ iff*

$$\nu^* = \{v \in \mathcal{V} | \inf_{v' \in \mathcal{V}} v'(\pi_V) \leq v(\pi_V) \leq \sup_{v' \in \mathcal{V}} v'(\pi_V)\}.$$

---
[1] We use the term "value" function to refer to the expectation of value throughout the processing of the diagram.

The next definition generalizes the idea of a *regular constraint* given in [2] to include value constraints:

**Definition 4 (Regular Constraint)** *A constraint is said to be a* regular constraint *iff it is equivalent to its regular extension.*

Based on these definitions, one can define functions which describe the regular constraints. For probabilities we have:

**Definition 5 (Constraint Function, Probability)** *The function $b_c(\mathbf{x})$ is said to be the* lower bound *of $c$ at the point $\mathbf{x}$ iff*

$$b_c(\mathbf{x}) = \inf_{p \in c} p(\mathbf{x}).$$

For value constraints, we have:

**Definition 6 (Constraint Functions, Value)** *If $\nu$ is a regular value constraint for a diagram $I$ then*

$$\nu_U(\pi_v) = \inf_{v' \in \mathcal{V}} v'(\pi_V)$$

$$\nu_L(\pi_v) = \sup_{v' \in \mathcal{V}} v'(\pi_V)$$

*are the* upper *and* lower value constraint functions *for $\nu$.*

In [1, 2], we found the need to refer to the concept of *compatibility* of a probability distribution with an influence diagram:

**Definition 7 (D-compatible, Probability)** *A joint distribution, $p$, is said to be* D-compatible *to an acyclic directed graph $I = \langle N, A \rangle$ if and only if there is a labelling of nodes in $\mathcal{U} = \{X_1, X_2, \ldots\}$ with associated variables $X^{(n)}$, such that*

$$p(x^{(1)}, x^{(2)}, x^{(3)} \ldots) = \prod_i p(x^{(i)} | s_{\Pi_{X^{(i)}}}) \quad (1)$$

Similarly, we have need of a compatibility concept for value functions with respect to an influence diagram:

**Definition 8 (D-compatible, Value)** *A value function for an influence diagram $I$ is* D-compatible *with $I$ with a value node $V$ if it is a function from the immediate predecessors of $V$ to the reals:*

$$v : \Pi_v \to \Re.$$

We now link these concepts with that of a regular constraint.

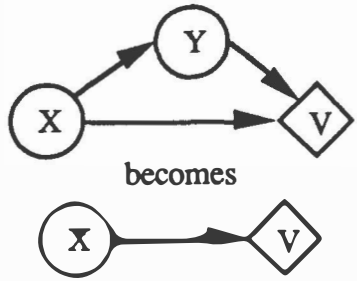

Figure 1: Removal of a chance node predecessor to the value node.

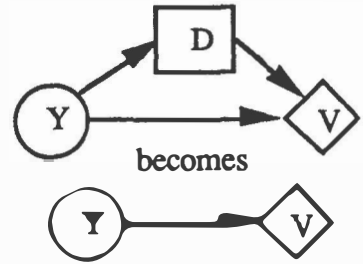

Figure 2: Removal of a decision node predecssor to the value node.

**Definition 9 (Diagram Regular, Value)** *We say a constraint $\nu$ is diagram regular with respect to a diagram $I$ iff for all $v \in \nu$ we have:*

1. *$v$ is D-compatible with $I$, and*
2. *$\nu$ is a regular constraint.*

**Definition 10 (Diagram Regular, Probability)** *We say a constraint $c$ is diagram regular with respect to a diagram $I$ iff there exists a set of regular constraints $c_{(i)}$ such that $p \in c$ iff:*

1. *$p$ is D-compatible with $I$, and*
2. *For each term in Equation 1,*

$$b_{c_{(i)}}(x^{(i)}|\pi_{X^{(i)}}) \leq p(x^{(i)}|\pi_{X^{(i)}})$$

*where $b_{c_{(i)}}$ is given by 5.*

## 3 Transformations

We now present three theorems which provide the fundamental operations necessary for evaluating an influence diagram to obtain a policy based on maximization of expected value [4]. A sequence of these operations (illustrated in Figures 2,2, and 4) are sufficient to evaluate any diagram [5].

**Theorem 1 (Chance Node Removal)**
*Consider a diagram $I$ with value node $V$ whose immediate predecessors are $X \in \mathcal{U} \cup \mathcal{D}$ and $Y \in \mathcal{U}$ and with $X$ an immediate predecessor of $Y$ (See Figure 1). Let $b_c(y|x)$ be the lower bound constraint function for $y$ given $x$ for a regular constraint $c$ and let $\nu_U(y,x)$ and $\nu_L(y,x)$ be the upper and lower value constraint functions for a diagram regular constraint $\nu$. Consider the single transformation on $I$ producing a new diagram with $Y$ removed. Then for all $x$,*

$$\nu_U(x) = \nu_U(y_r, x)u_c(y_r|x) + \sum_{i \neq r} \nu_U(y_i, x)b_c(y_i|x)$$

$$\nu_L(x) = \nu_L(y_s, x)u_c(y_s|x) + \sum_{i \neq s} \nu_U(y_i, x)b_c(y_i|x),$$

*where $y_s$ and $y_r$ depend on $x$ and are such that*

$$\forall i, \nu_U(y_r, x) \geq \nu_U(y_i, x)$$

$$\forall i, \nu_L(y_s, x) \leq \nu_L(y_i, x),$$

*are the least upper bound and the greatest lower bound respectively for $v(x) = \sum_y v(y,x)p(y|x)$ over all $p(\cdot|\cdot) \in c$ and $v(\cdot, \cdot) \in \nu$.*

Theorem 1 provides a method of calculating new intervals for the value node given an initial set of intervals for the value node and the chance node predecessor to be removed. The proof to this theorem is an almost immediate consequence of Lemma 1 in [2].

**Theorem 2 (Decision Node Removal)**
*Consider a diagram $I$ with value node $V$ having immediate predecessors $Y \in \mathcal{U} \cup \mathcal{D}$ and decision node $D \in \mathcal{D}$, with $Y$ an immediate predecessor of $D$ (See Figure 2). Let $\nu_U(y,d)$ and $\nu_L(y,d)$ be the upper and lower value constraint functions for a regular constraint $\nu$. Consider the single transformation on $I$ producing a new diagram with $D$ removed. Then for all $y$,*

$$\nu_U(y) = \max_{d \in S(y)} \nu_U(y,d) \quad (2)$$

$$\nu_L(y) = \min_{d \in S(y)} \nu_L(y,d) \quad (3)$$

$$\text{with } S(y) = \{d_l | \neg(\exists j, \nu_U(y,d_l) < \nu_L(y,d_j)\} \quad (4)$$

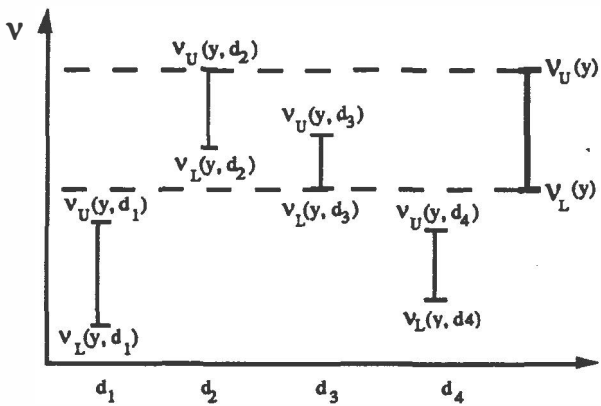

Figure 3: Determination of new value intervals from value intervals associated with individual decision alternatives. Alternative $d_2$ dominated by $d_1$ and $d_4$, but not $d_3$. The new interval is determined from the bounds imposed by the non-dominated alternatives.

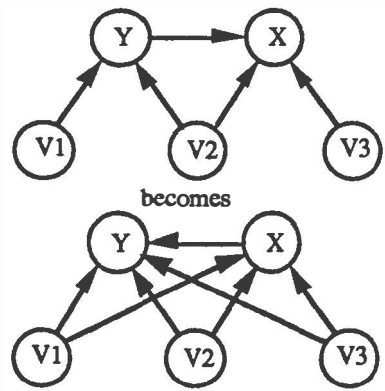

Figure 4: Reversal of an arc in an influence diagram.

*are the least upper bound and the greatest lower bound respectively for*

$$v(y) = v(y, d^*(y))$$

*for all $v(y,d) \in \nu$, where $d^*(y)$ is the optimal decision policy for value $v$, solving $\max_d v(y,d)$.*

Theorem 2 provides a method of calculating new intervals for the value node given an initial set of intervals for the value node.[2] Equation(s) 2 (3) says that the new upper (lower) bound in value is just the maximum (minimum) of the previous upper (lower) bounds on value, among the *admissible* decision alternatives. Admissibility is defined in Equation 4. An alternative $d_l$ is admissible if there does not exist an alternative whose value interval strictly dominates. This notion and the calculation of new value intervals is illustrated in Figure 3.

In lieu of the single decision policy recommendation generated by a point-valued influence diagram, the interval-valued procedure creates the sets $S(y)$, which define the admissible decisions given values for the predecessors of the decision node. We have the following simple corollary with regard to the admissible set, $S(y)$:

**Corollary 1 (Admissibility)** *Consider the same conditions as in Theorem 2. Let*

$$S(y) = \{d_l | \neg(\exists j, \nu_U(y, d_l) < \nu_L(y, d_j))\}.$$

---
[2]The proof is straightforward, and brevity is omitted.

*Let $v$ be any value function in $\mathcal{V}$. Let $d^*(y)$ denote the optimal decision policy for this value function. Then, for all $y$, $d^*(y) \in S(y)$.*

This corollary states that the optimal policy which would have been generated by the point-valued procedure is included in the set of admissible decisions generated by the interval-valued procedure.

Loui in [3] defines two separate criteria for admissibility with interval valued probabilities. The first is as stated above. The second (paraphrased) is that $d$ is "E-admissible" iff

$$\exists p \in c, v \in \nu \ni \forall d_i,$$

$$\sum_y v(y,d) p(y,d) \geq \sum_y v(y,d_i) p(y,d_i).$$

It is fairly straightforward to show that, in the case of diagram regular constraints, these two definitions are equivalent. Finally, we state the reversal theorem given in [2], generalizing it slightly to include decision nodes as possible predecessors.

**Theorem 3 (Reversal)** *Consider a diagram $I$ with chance node $X \in \mathcal{U}$ immediate predecessor $Y \in \mathcal{U}$, with $V_1 \in \mathcal{U} \cup \mathcal{D}$ a predecessor of $Y$ but not $X$, $V_2 \in \mathcal{U} \cup \mathcal{D}$ a common predecessor of $X$ and $Y$, and $V_3 \in \mathcal{U} \cup \mathcal{D}$ a predecessor of $X$ but not $Y$. (See Figure 4). Given lower bound constraint functions $b_c(x|y, v_2, v_3)$ and $b_d(y|v_1, v_2)$ for all values of $x$ and $y$ and associated regular constraint sets $c$ and $d$, together with their corresponding upper-constraint functions $u_c(x|y, v_2, v_3)$ and $u_d(y|v_1, v_2)$. Consider the single transformation to the diagram reversing the direction of the arc between $X$ and $Y$. Let*

$$W(x, y, v_1, v_2, v_3) \qquad =$$

$$b_c(x|y, v_2, v_3)b_d(y|v_1, v_2)$$
$$+ u_c(x|y_s, v_2, v_3)u_d(y_s)$$
$$+ \sum_{y_i \neq y_s, y} u_c(x|y_i, v_2, v_3)b_d(y_i|v_1, v_2)$$

with $y_s$ chosen such that $y_s \neq y$ and $u_c(x|y_s, v_2, v_3) \geq u_c(x|y_i, v_2, v_3)$ for all $y_i \neq y_s, y$. For all $x$ and $y$, we define $b^*(y|x, v_1, v_2, v_3)$ as follows:

1. If $W(x, y, v_1, v_2, v_3) > 0$ then

$$b^*(y|x, v_1, v_2, v_3) = \frac{b_c(x|y, v_2, v_3)b_d(y|v_1, v_2)}{W(x, y, v_1, v_2, v_3)}, \quad (5)$$

2. If $W(x, y, v_1, v_2, v_3) = 0$, and there exists $y_j \neq y, y_s$ such that $u_c(x|y_j, v_2, v_3)u_d(y_j|v_1, v_2) > 0$, then we take by convention:

$$b^*(y|x, v_1, v_2, v_3) = 0.$$

3. Otherwise $b^*(y|x, v_1, v_2, v_3)$ is indeterminate.

When $b^*$ is determined, it is the greatest lower bound for

$$p(y|x, v_1, v_2, v_3) = \frac{p(x|y, v_2, v_3)p(y|v_1, v_2)}{\sum_y p(x|y, v_2, v_3)p(y|v_1, v_2)},$$

with $p(x|y, v_2, v_3) \in c$ and $p(y|v_1, v_2) \in d$.

## 4 Soundness and Optimality

In [2] we proved soundness and optimality properties for transformations to interval-valued probability networks. In this section we state the analogous theorems for diagrams which include decision and value nodes and use the transformations stated in the previous section.

Referring to Figure 5, we define:

**Definition 11 (Soundness)** *Let $H$ be a single operation on a diagram $I$ to produce a new diagram $I' = H(I)$. $H$ represents either an arc reversal or a node removal. We define any interval transformation algorithm, $\mathcal{A}$, as an operation on diagram regular constraint $c_I$ and diagram regular value constraint $\nu_I$ to produce corresponding diagram regular constraints $\mathcal{A}_H(c_I)$ and $\mathcal{A}_H(\nu_I)$ for $I'$. We say $\mathcal{A}$ is sound if for all $p \in c_I$ and all $v \in \nu_I$ we have $\overline{H}(p) \in \mathcal{A}_H(c_I)$ and $\overline{H}(v) \in \mathcal{A}_H(\nu_I)$, where $\overline{H}$ is the transformation on distributions and value functions D-compatible with $I$ to distributions and value functions D-compatible with $I'$, $\mathcal{A}_H(c_I)$ is the set of probability distributions produced by $\mathcal{A}$ for $I'$ that corresponds to the operation $H$.*

This definition says that a transformation is sound if the new value function and probability distribution that one would have obtained by applying the exact transformation to individual members of the original constraint sets is contained in the sets produced by the operations described in Theorems 1-3. We state without proof:

**Theorem 4 (Soundness)** *Algorithm $\mathcal{A}^o$, consisting of a node removal or arc reversal as detailed in Theorems 1, 2, and 3, is sound.*

Soundness is a weak condition. We need to show that the intervals calculated by Theorems 1-3 are best in some appropriate sense.

**Theorem 5 (Minimality)** *Let $I = \langle N, A \rangle$ be an influence diagram and let $c_I$ and $\nu_I$ be diagram regular constraints on the probabilities and value functions for $I$. Let $H$ be a single topological operation on $I$ producing $I'$, let $\mathcal{A}_H^o(c_I)$ and $\mathcal{A}_H^o(\nu_I)$ be the constraints on the distributions and value functions D-compatible with $I'$ produced by $H$. Let $\overline{H}$ be the mapping corresponding to $H$ from distributions and value functions D-compatible with $I$ to those D-compatible with $I'$. Then, letting $\mathcal{C}(I')$ and $\mathcal{L}(I')$ denote the set of diagram regular constraints for probability functions and value functions with respect to the image diagram $I'$ and if*

$$\mathcal{B}' = \{c_{I'}|\overline{H}(c_I) \subseteq c_{I'}, \text{ and } c_{I'} \in \mathcal{C}(I')\},$$
$$\mathcal{M}' = \{\nu_{I'}|\overline{H}(\nu_I) \subseteq \nu_{I'}, \text{ and } \nu_{I'} \in \mathcal{L}(I')\},$$

*we have,*

$$\mathcal{A}_H^o(c_I) = \inf_{c_{I'} \in \mathcal{B}'} c_{I'} = \bigcap_{c_{I'} \in \mathcal{B}'} c_{I'}.$$

$$\mathcal{A}_H^o(\nu_I) = \inf_{\nu_{I'} \in \mathcal{M}'} \nu_{I'} = \bigcap_{\nu_{I'} \in \mathcal{M}'} \nu_{I'}.$$

This theorem says that each transformation on constraints given by algorithm $\mathcal{A}^o$ produces a diagram regular constraint set which is the smallest of all such sets that remain sound.[3]

## 5 Empirical Results

The approach described in this paper has been implemented and tested on a variety of influence diagrams. In this section we describe some experiments on a particular diagram and illustrate use of the algorithm to examine robustness and sensitivity of results. For the purposes of this discussion, we have encoded the oil-wildcatter's decision model as an influence diagram (See Figure 6). The model has two decisions. The node labelled TEST is the choice among alternative geologic tests of the seismic structure in an area. No test, a cheap test and a perfect test are the alternatives. The other decision is whether or not to drill. The arcs into the DRILL node indicate that the type of test and its result will be available when deciding whether or not to drill. TEST-RESULTS provides information about SEISMIC-STRUCTURE, which in turn provides information about the AMOUNT-OF-OIL.

---
[3]For brevity the proof is omitted.

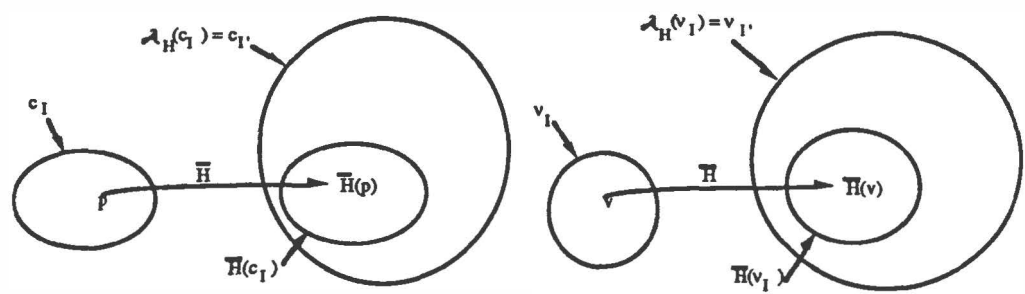

Figure 5: Mappings on probability distributions. $H$ corresponds to a topological operation on a diagram $I$ to produce a new diagram $I' = H(I)$. $\overline{H}$ represents a mapping from the space of probability distributions or value functions that are D-compatible with $I$ to the corresponding spaces that are D-compatible with $I'$. An interval transformation algorithm $\mathcal{A}$ maps constraint sets into constraint sets. The algorithm is *sound* if $\overline{H}(c_I) \subseteq \mathcal{A}_H(c_I)$ and $\overline{H}(\nu_I) \subseteq \mathcal{A}_H(\nu_I)$.

One way to characterize the interval influence diagram (IID) approach described here is to compare it to an exact approach to calculating the ramifications of interval inputs. The results labelled EXACT below refer to calculating values and decision recommendations for all combinations of the endpoints of the input probability ranges. Tables 1 and 2 display the impact of using a lower-bound inerval approach when three different levels of imprecision are added to the original diagram. Specifically, we examined probability ranges[4] of .01, .05 and .10 for three nodes (AMOUNT-OF-OIL, SEISMIC-STRUCTURE, and COST-OF-DRILLING). The exact procedure consisted of solving the network for optimal decisions for each of 1296 possible configurations. The exact expected value ranges and admissible decision sets were based on these runs.

---
[4] The range $R = u(x_i) - b(x_i) = 1 - \sum_j b(x_j)$. For this study, we selected a subset of nodes in the diagram for analysis and introduced this range into each distribution residing in the node.

The primary decision is whether or not to DRILL. Recall from the influence diagram that the DRILL decision is conditioned on the type of test selected and its result (one of "NS", "OS", or "CS"). Tabel 1 shows the admissible decisions for the the various possible information states for the DRILL decision, using interval influence diagrams and the exact procedure. The IID and the EXACT procedures provide identical sets of admissible decisions for this variable, indicating for this decision IID is a perfect approximation to the exact analysis in terms of decision recommendations.

For the TEST decision IID is a less than perfect approximation. Table 2 shows that as soon as any imprecision is introduced, the IID procedure is unable to distinguish among the alternatives for TEST. At a .10 level of imprecision, the EXACT algorithm cannot distinguish between the "none" and "cheap" test options. The table also shows the intervals in expected value associated with each procedure at each level of imprecision.

We can also use the IID procedure to explore the sensitivity of results to imprecision in various sets of chance nodes. For example, in Table 3 admissible decision sets for the TEST decision have been generated far probability range .05 for various subsets of the nodes AMOUNT-OF-OIL, SEISMIC-STRUCTURE, and COST-OF-DRILLING. The table indicates that results are least sensitive to imprecision in conditional probabilities for COST-OF-DRILLING. Sensitivity of results to SEISMIC-STRUCTURE and AMOUNT-OF-OIL are approximately equivalent according to the table.

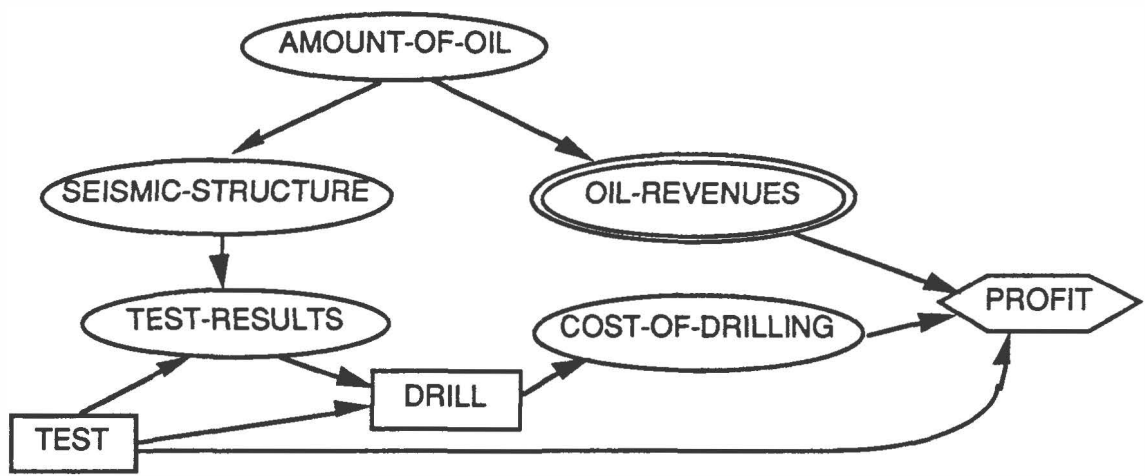

Figure 6: The influence diagram for the oil-wildcatter. Profits depend on OIL-REVENUES, COST-OF-DRILLING (depending on whether the DRILL decision), and TEST. TEST represents a choice among geologic tests whose results are available when the DRILL decision is made.

| Range | EXACT | | | IID | | |
|---|---|---|---|---|---|---|
| | $\nu_L$ | $\nu_U$ | TEST type | $\nu_L$ | $\nu_U$ | TEST type |
| .00 | 40.0 | 40.0 | none | 40.0 | 40.0 | none |
| .01 | 38.9 | 42.3 | none | 30.0 | 43.8 | none/cheap/perfect |
| .05 | 34.5 | 51.5 | none | 8.0 | 69.4 | none/cheap/perfect |
| .10 | 29.0 | 63.0 | none/cheap | -9.2 | 99.8 | none/cheap/perfect |

Table 2: The table shows the sets of admissible decisions for the TEST decision for various input probability ranges.

| Nodes Analyzed | Range = .05 | | | Range = .01 | | |
|---|---|---|---|---|---|---|
| | $\nu_L$ | $\nu_U$ | TEST type | $\nu_L$ | $\nu_U$ | TEST type |
| AMOUNT-OF-OIL SEISMIC-STRUCTURE COST-OF-DRILLING | 8.0 | 69.4 | none/cheap/perfect | 30.0 | 43.8 | none/cheap/perfect |
| SEISMIC-STRUCTURE COST-OF-DRILLING | 14.6 | 57.1 | none/cheap/perfect | 34.3 | 41.1 | none/cheap |
| AMOUNT-OF-OIL COST-OF-DRILLING | 21.1 | 53.0 | none/cheap/perfect | 35.3 | 42.3 | none/cheap/perfect |
| AMOUNT-OF-OIL SEISMIC-STRUCTURE | 9.0 | 66.9 | none/cheap/perfect | 30.1 | 43.3 | none/cheap/perfect |
| AMOUNT-OF-OIL | 27.3 | 50.5 | none/cheap/perfect | 35.5 | 41.8 | none/cheap |
| SEISMIC-STRUCTURE | 15.6 | 54.6 | none/cheap/perfect | 34.5 | 40.6 | none/cheap |
| COST-OF-DRILLING | 36.0 | 42.5 | none/cheap | 39.8 | 40.5 | none |

Table 3: The table shows the sets of admissible decisions for the DRILL decision for various input probability ranges.

| Range | TEST | TEST RESULT | DRILL? EXACT | DRILL? IID |
|---|---|---|---|---|
| .01 | none | none | yes | yes |
| | cheap | NS | yes | yes |
| | | OS | yes | yes |
| | | CS | yes | yes |
| | perfect | NS | no | no |
| | | OS | yes | yes |
| | | CS | yes | yes |
| .05 and .10 | none | none | yes | yes |
| | cheap | NS | yes/no | yes/no |
| | | OS | yes | yes |
| | | CS | yes | yes |
| | perfect | NS | yes/no | yes/no |
| | | OS | yes | yes |
| | | CS | yes | yes |

Table 1: The table shows the sets of admissible decisions for the DRILL decision for various input probability ranges.

## 6 Conclusions

In this paper we have extended previous results in interval values for influence diagrams to include decision making. While manipulation of belief has many interesting technical properties, the importance of varying the degree of precision in probabilities can only be gauged by including values and decisions into the analysis. This paper represents one step in that direction.

## References

[1] K.W. Fertig and J.S. Breese. Interval influence diagrams. In *Proceedings of Fifth Workshop on Uncertainty in Artificial Intelligence*, Detroit, MI, August 1989.

[2] K.W. Fertig and J.S. Breese. Probability intervals over influence diagrams. Technical report, Rockwell International Science Center, March 1990. Rockwell Research Report 4.

[3] R. Loui. *Theory and Computation of Uncertain Inference*. PhD thesis, Department of Computer Science, University of Rochester, 1987. Also available as TR-228, University of Rochester, Department of Computer Science.

[4] S.M. Olmsted. *On Representing and Solving Decision Problems*. PhD thesis, Department of Engineering-Economic Systems, Stanford University, December 1983.

[5] R.D. Shachter. Evaluating influence diagrams. *Operations Research*, 34:871–882, 1986.